\documentclass[10pt,twocolumn,letterpaper]{article}

\usepackage{cvpr} 
\definecolor{MyDarkBlue}{rgb}{0,0.08,1}
\definecolor{MyDarkGreen}{rgb}{0.02,0.6,0.02}
\definecolor{MyDarkRed}{rgb}{0.8,0.02,0.02}
\definecolor{MyDarkOrange}{rgb}{0.40,0.2,0.02}
\definecolor{MyPurple}{RGB}{111,0,255}
\definecolor{MyRed}{rgb}{1.0,0.0,0.0}
\definecolor{MyGold}{rgb}{0.75,0.6,0.12}
\definecolor{MyDarkgray}{rgb}{0.66, 0.66, 0.66}
\definecolor{MyLightGray}{rgb}{0.8, 0.8, 0.8}
\definecolor{MyWineRed}{rgb}{0.694,0.071, 0.149}
\definecolor{nicegreen}{rgb}{0.1, 0.6, 0.2}

\usepackage{pgfplots}
\usetikzlibrary{spy}
\usetikzlibrary{calc}
\usepackage{tikz,pgfplots,pgfplotstable}
\definecolor{col1}{rgb}{0.04314, 0.51765, 0.64706}
\definecolor{cvprblue}{rgb}{0.21,0.49,0.74}
\usepackage[pagebackref,breaklinks,colorlinks,allcolors=cvprblue]{hyperref}

\def\paperID{1339} % *** Enter the Paper ID here
\def\confName{CVPR}
\def\confYear{2025}

\definecolor{peach}{rgb}{ 0.943, 0.188, 0.526}
\definecolor{plum}{rgb}{ 0.858, 0.188, 0.478}
\definecolor{muted_navy_blue}{RGB}{63, 75, 166}
\definecolor{muted_sky_blue}{RGB}{134,166,213}
\definecolor{federal_blue}{RGB}{0,96,240}
\definecolor{regulation_red}{RGB}{226, 20, 79}
\definecolor{federal_gold}{RGB}{240, 212, 14}

\usepackage{subcaption}
\usepackage{wrapfig}
\usepackage{enumitem}
\usepackage{graphicx}
\usepackage{amsmath}
\usepackage{booktabs}
\usepackage{enumitem}

\usepackage{bm}
\usepackage{amsmath}
\usepackage{colortbl}
\usepackage{multirow}
\usepackage{tikz}
\usepackage{wrapfig}
\usepackage{cases}
\newcommand{\mytilde}{\raise.17ex\hbox{$\scriptstyle\mathtt{\sim}$}}
\usepackage{algpseudocode}
\usepackage[ruled,vlined]{algorithm2e} % added it in commands.tex under Files :) 
\usepackage{arydshln}

\title{Quaffure: Real-Time Quasi-Static Neural Hair Simulation}

\author{Tuur Stuyck$^{*}$ \hspace{0.3em} Gene Wei-Chin Lin$^{*}$ \hspace{0.3em} Egor Larionov \hspace{0.3em} Hsiao-yu Chen \\ Aljaz Bozic \hspace{0.3em} Nikolaos Sarafianos \hspace{0.3em} Doug Roble \\
{\normalsize Meta Reality Labs} \\
{\normalsize $^*$equal contributions} 
}
\begin{document}
\twocolumn[{%
\renewcommand\twocolumn[1][]{#1}%
\maketitle
\vspace{-0.7cm}
\centering
    \includegraphics[width=0.93\textwidth]{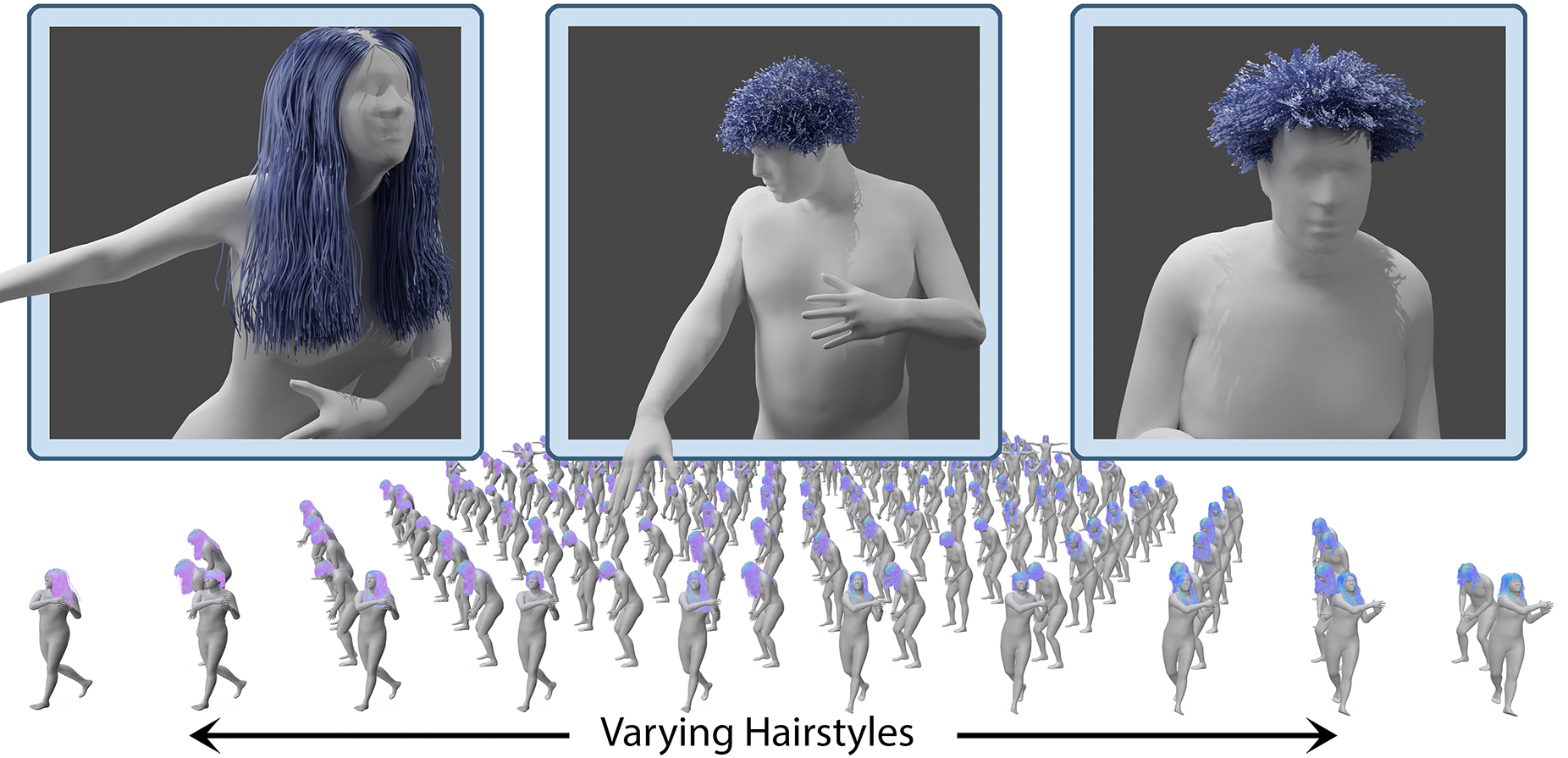}
    \captionof{figure}{We present \emph{Quaffure}, a real-time quasi-static neural hair simulator, which produces naturally draped hair in only a few milliseconds on commodity hardware, taking the hairstyle, body shape and pose into account. Our method scales to predicting the drape of 1000 hair grooms in just 0.3 seconds. Quaffure is trained using a physics-based self-supervised loss, eliminating the need for simulated training data that is costly and cumbersome to obtain. We show that our method works for a wide variety of body shapes and poses with a range of hairstyles varying from straight to curly, short to long.}
    \vspace*{0.25cm}
    \label{fig:teaser}
}]

\begin{abstract}
Realistic hair motion is crucial for high-quality avatars, but it is often limited by the computational resources available for real-time applications. To address this challenge, we propose a novel neural approach to predict physically plausible hair deformations that generalizes to various body poses, shapes, and hairstyles. 
Our model is trained using a self-supervised loss, eliminating the need for expensive data generation and storage. We demonstrate our method's effectiveness through numerous results across a wide range of pose and shape variations, showcasing its robust generalization capabilities and temporally smooth results. 
Our approach is highly suitable for real-time applications with an inference time of only a few milliseconds on consumer hardware and its ability to scale to predicting the drape of 1000 grooms in 0.3 seconds. 
% Code will be released.
\end{abstract}

\vspace{-0.3cm}
\section{Introduction}
Hair is a crucial aspect of realistic avatars, playing a vital role in real-time applications such as games and telepresence, as well as offline generation of digital characters for film. Over the past decades, computer graphics research has devoted significant attention to the topic of hair, in order to create believable and engaging virtual characters. Hair modeling is a complex process that involves multiple stages, including grooming, simulation, and rendering. Previous research has explored various approaches to hair motion modeling, including physics-based simulations \cite{realTimeHairMeshesSimulation, daviet2023interactive, huang2023towards} and data-driven methods \cite{wang2022neuwigs, guan2012multi, Wang_2022_CVPR}. However, hair remains a challenging phenomenon to model due to the high number of degrees of freedom required to represent a full head of hair, which typically consists of hundreds of thousands of strands.
To make hair motion modeling more tractable, it is common practice to model a full groom with a limited number of strands, often referred to as guide strands. This approach involves modeling a subset of strands to represent the full groom, and then recovering the full resolution hair groom from these guide strands for rendering \cite{hsu2024real}.

Simulating the dynamics of hair motions has received significant attention in the field. Several approaches have been developed to model the behavior of hair strands ranging from efficient mass-spring systems~\cite{selle2008mass} to rod models~\cite{bergou2008discrete, sca2016cosserat, cgf2018cosserat} able to model bending and twisting, producing complex coiling motions to volumetric approaches \cite{bando2003animating, mcadams2009detail, petrovic2005volumetric}. Despite impressive visual results, these models still require large computational resources making them ill-suited for real-time applications. Many advances in the field of physics-based simulation has resulted in efficient algorithms that are able to exploit parallelism of GPU hardware, resulting in real-time simulations~\cite{bouaziz2023projective, muller2007position, macklin2016xpbd}. However, these still require dedicated hardware resources \cite{daviet2023interactive, realTimeHairMeshesSimulation} and compute time might vary based on the resolution and number of the hair strands as well as the number of collisions occurring in the scene. Additionally, physics-based simulation can often be unstable and require significant parameter tuning, often leading to unpredictable or undesirable results. To address this, subspace simulation has been applied to hair dynamics but this remains limited to models that overfit to a single groom only~\cite{chai2014reduced}. Learned simulators~\cite{lyu2020real, zhou2023groomgen} have shown promise for modeling real-time hair dynamics but require large data sets, which are cumbersome to obtain. In general, supervised learning requires training data, which needs to be generated using physics-based simulation. This is both expensive in terms of time and storage required but also requires expert knowledge and access to specialized tools to produce. There is no publicly available data set that could be leveraged.

In this paper, we propose a novel and efficient neural hair simulation approach to model hair quasi-statics. Our method leverages a learning-based method to predict physically plausible hair deformations, while also generalizing to various body poses and shapes. We opted for training in a self-supervised manner, eliminating the need for expensive data generation and storage. Our approach achieves an inference performance of just a few milliseconds on standard consumer hardware, without requiring elaborate code optimizations or specialized hardware features. 
In contrast to physics-based simulation methods, the inference time of our approach is fixed and is independent of the number of strands in the groom or number of collisions in the scene. Our results are stable and the method scales to predicting hundreds of grooms simultaneously at interactive rates.
In summary, our contributions are 
\begin{itemize}
    \item A hair decomposition strategy consisting of a hair pose module and a learned corrector, which enables the use of an efficient neural decoder able to produce quasi-static hair results in milliseconds and scales to running a thousand grooms at interactive rates. 
    \item We are the first to propose a self-supervised learning approach for hair that completely eliminates the necessity of computing and storing large amounts training data, which would require expert knowledge and access to specialized tools to produce.
    \item A modified formulation of the Cosserat energy for modeling hair. Our formulation is efficient in its evaluation and we demonstrate that it is well-suited for neural network training. Our proposed model is able to preserve hair strand shape for a wide range of hairstyles including very curly hair.
    \item We demonstrate how a single trained network efficiently models different hairstyles, generalizing to body poses and shapes.
\end{itemize}

\begin{figure*}[t]
    \centering
    \includegraphics[width=0.99\linewidth]{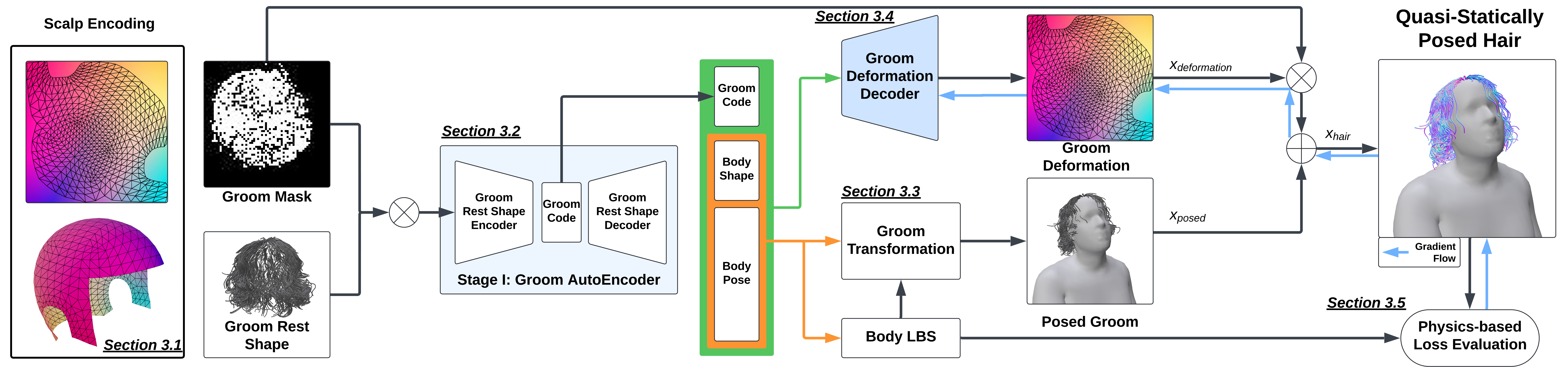}
    \caption{\textbf{Quaffure Overview}: Our method takes a code as input, consisting of a latent code for the rest hair shape, body shape parameters, and full skeleton pose. 
    The output is naturally draped hair produced as the sum of posed hair given the body pose and shape parameters, combined with learned corrections which are produced by the groom deformation decoder. 
    We train our method in two stages: i) an autoencoder is trained on all hairstyles to obtain a groom latent code, and ii) the groom deformation decoder is trained in a physics-based self-supervised fashion. The hair strands are encoded in a 2D texture representation (left) where strands are encoded in the pixel in which the root particle is located. The figure shows how the 3D scalp geometry (bottom) is mapped to a high dimensional 2D texture map (top).}
    \label{fig:pipeline}
    \vspace{-0.15cm}
\end{figure*}

\section{Related work}
\textbf{Strand-based Hair Reconstruction.}
In recent years there has been significant progress in hair reconstruction research. Recovering hair strands has been especially sought after, since hair reconstructed as strands can easily be integrated into existing game engines, to be animated and rendered with physics-based simulation and rendering. Neural Strands~\cite{rosu2022neuralstrands} represent the hair groom as a neural texture on the head scalp, where each texel feature can be decoded into an explicit strand. 
This representation was later trained on a dataset of synthetic grooms, to provide prior for strand-based hair fitting to multi-view images \cite{sklyarova2023neural}.
\citet{zhou2024groomcap} proposes a prior-free reconstruction approach. It was extended to support text-guided hair generation in HAAR \cite{sklyarova2023haar}. The fitting quality was further improved in \cite{luo2024gaussianhair, zakharov2024gh, xu2023gaussianheadavatar} by additionally decoding Gaussian splats~\cite{kerbl20233d} along the strands, representing hair appearance. Instead of requiring multi-view capture, several works explored strand-based hair reconstruction from a dynamic monocular video~\cite{yang2019dynamic} or just a single image~\cite{wu2022neuralhdhair, zhou2018hairnet}, where volumetric 3D hair orientations were used to model hair geometry.  To achieve the highest reconstruction quality, \citet{shen2023ct2hair} captured volumetric CT scans to reconstruct strand-based grooms.

\noindent\textbf{Learning-based Hair Models.}
To either accelerate or improve the realism of physics-based simulation or rendering methods, many learning-based hair models have been introduced. \citet{Wang_2022_CVPR} use a hybrid representation, with volumetric primitives aligned along guide strands to model hair appearance under motion. To simplify fitting to real observations, \citet{wang2022neuwigs} used a mixture of volumetric primitives that jointly represent hair geometry and appearance. A unified latent representation for diverse hairstyles has been explored in \cite{zhou2023groomgen, he2024perm}, where explicit strand geometry is encoded into a latent space using a variational autoencoder (VAE) \cite{zhou2023groomgen} or a generative adversarial network (GAN) \cite{he2024perm}, enabling hair generation and interpolation across hairstyles. To accelerate hair simulation, \citet{guan2012multi} proposed a multi-linear model for representing hair dynamics, and \citet{lyu2020real} introduced a neural upsampler to efficiently convert a low number of guide strands into dense hair.

\noindent\textbf{Physics-based Hair Models.}
Realistic hair simulation~\cite{HWU2024, WUSHI2023} has been an active area of research in computer graphics for decades. To achieve high-fidelity results, it is necessary to simulate every individual strand and their interactions with each other. Common elastic models used for strand-based hair include mass-spring models~\cite{selle2008mass} and elastic rods~\cite{bergou2008discrete, sca2016cosserat, cgf2018cosserat}. In addition, complex hair-hair interactions such as self-collision~\cite{mcadams2009detail} and friction forces~\cite{kaufman2014adaptive} are also important for realistic behavior.
Several reduced models have been developed to lower the computational budget for real-time hair simulation. One idea is to simulate the hair as layered elastic bodies~\cite{yuksel2009hair, bhokare2024real}, while another suggests representing a group of hairs as strips~\cite{koh2001simple, koh2000real}. The most popular approach is to simulate a subset of the overall hair as guide strands and interpolate between the guides for individual strand detail~\cite{chai2014reduced, hsu2024real}. \citet{hu2017simulation} propose a method to automatically determine simulation parameters from video footage.

\noindent\textbf{Physics-based Self-supervised Learning.}
Data-driven methods have been successful in the field of simulation~\cite{raissi2019physics, fulton2019latent, zehnder2021ntopo}. However, data-driven methods do not ensure that the simulated system adheres to physical constraints, which can result in non-physical states. To address this, researchers have proposed incorporating physics energy loss directly into the training process to eliminate the need for generating large amounts of data and to enable better generalization. Physics-based self-supervised methods have shown promise in learning a reduced latent space for quasi-static states of rigid motion~\cite{sharp2023data} and cloth drapes~\cite{de2023drapenet}, which has been expanded to dynamic motions of cloth~\cite{santesteban2022snug, bertiche2022neural, grigorev2023hood, kair2024neuralclothsim}. However, there is limited research on hair simulation. The neural simulator of GroomGen~\cite{zhou2023groomgen} only demonstrated hair deformation for varying gravity directions, while our method is the first to apply a physics-supervised method for hair that generalizes to full motions and incorporates collision under different poses and body shapes.

\section{Methodology}
Our approach shown in Figure~\ref{fig:pipeline}, enables real-time estimation of posed quasi-static hair $\textbf{x}_\text{hair}$ based on hairstyle, body shape and pose. We achieve this by splitting the problem at hand in two parts. 
First, a simple and efficient groom transformation model transforms the hair groom rigidly with body pose and shape variations expressed as $\textbf{x}_\text{posed}$. Second, to model the groom specific pose and shape dependent variations $\textbf{x}_\text{deformation}$, we train a network to produce deformation fields, which when added to the posed hair, result in a physically plausible quasi-static drape that takes physical effects such as gravity and collisions into account. 
The final hair drape is simply obtained as $\textbf{x}_\text{hair} = \textbf{x}_\text{posed} + \textbf{x}_\text{deformation}$. 

\subsection{Representation}

We encode all hair geometry into a texture based representation based on the UV map of the scalp geometry as shown in 
the left of Figure~\ref{fig:pipeline}. Where a hair strand root particle falls into a pixel in the map, we encode a high dimensional feature, which represents the strand geometry. For a texture with height and width T and a groom with N vertices per strand, the encoding is of dimension $\mathbb{R}^{T \times T \times N \times 3}$.

\noindent \textbf{Body Model}: Our body model consists of body geometry rigged with a skeleton model, which enables the body geometry to change pose using linear blend skinning. 
The skeleton pose is represented using a set of pose parameters $\mathbf{\theta} \in \mathbb{R}^{81}$. To model variations in body shape, we leverage a statistical body model. This model has been constructed from a large data set of human body scans, to which our body geometry has been registered. Body shape variations are modeled as variations on top of an average shape and are parameterized using body shape coefficients $\mathbf{\beta} \in \mathbb{R}^{10}$.

\subsection{Groom Autoencoder}
We leverage an autoencoder to obtain a compact representation for all the different grooms in their rest shape. This model is trained separately and produces compact latent codes in $\mathbb{R}^{16}$ to distinguish different hairstyles. The autoencoder input and output are grooms encoded on the scalp as high dimensional texture maps. 
The autoencoder structure is made up of 2D convolutional down and up sampling layers for the encoder and decoder respectively.
Once the groom autoencoder is fully-trained, we discard its decoder since we are only interested in the latent groom codes produced by the encoder. The weights of the encoder are frozen for the follow-up training of the groom deformation decoder discussed in Section~\ref{ssec:gdd}. 

\subsection{Pose-based Groom Transformation}
\begin{figure}[t]
    \centering
    \includegraphics[width=1.0\linewidth]{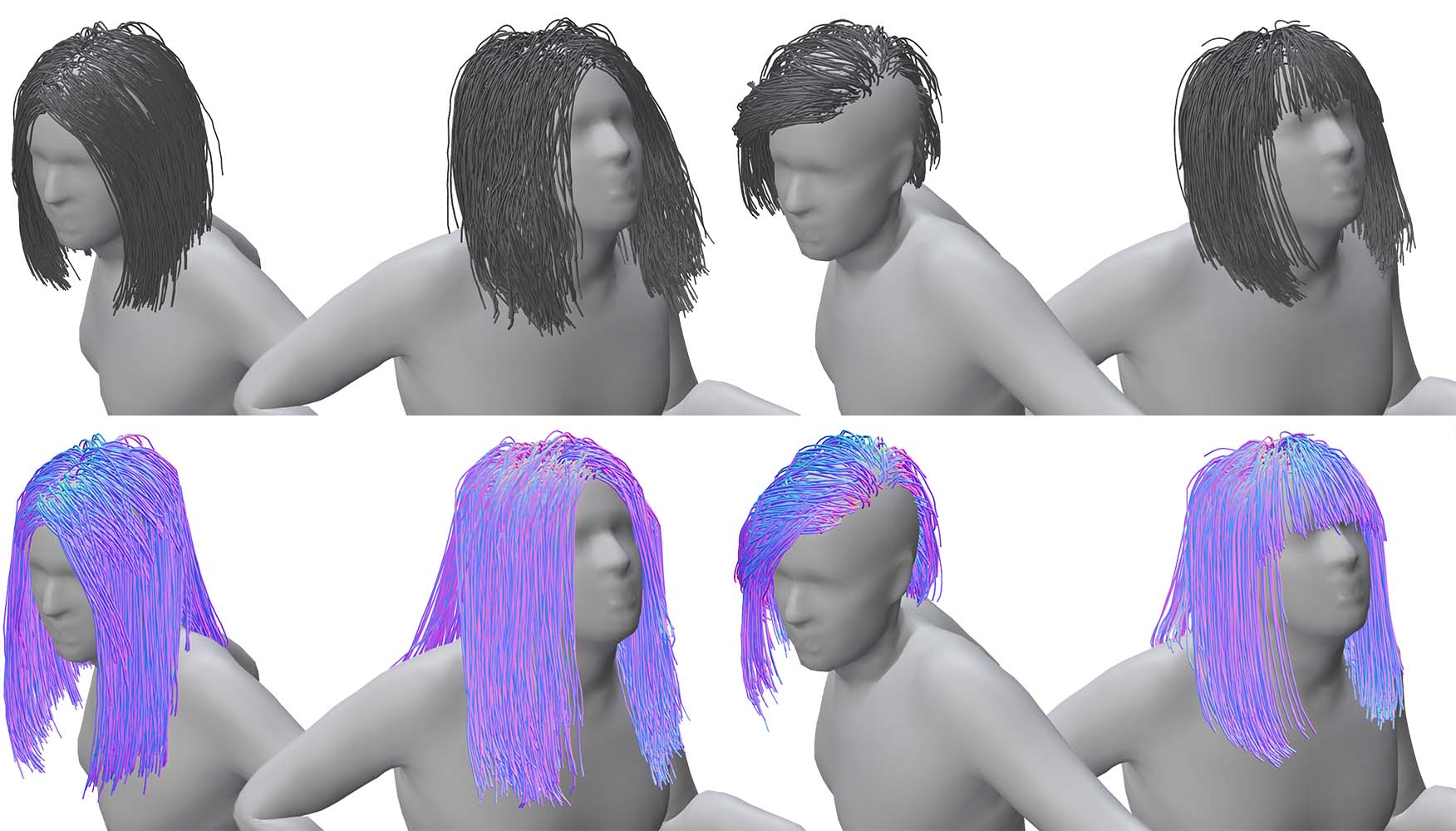}
    \caption{\textbf{Pose \& Shape-based Deformations}: Example of posed groom (top) which accounts for rigid rotations and translations of the rest shape only and the combined posed groom with learned deformations (bottom), which accounts for physical effects such as strand material model, gravity and collisions. 
    Despite body intersections incurred by the rigid transformation applied by the groom transformation module, our proposed network resolves all collisions with the body after applying the learned deformations (bottom). Note the significant change resulting in a natural drape.}
    \label{fig:lbsandoffset}
    \vspace{-0.2cm}
\end{figure}

Our pose-based transformation module rigidly transforms the groom in rest shape to account for any rotation and translation changes of the head motion based on the skeleton pose. Every individual hair strand is attached to a scalp triangle at the root. Given the rest configuration, we compute a local coordinate system based on the triangle edges and compute embedding of the hair strand vertices in this local axis. This allows us to efficiently update the strand geometry to move rigidly with the head motion and shape deformations. The top row of Figure~\ref{fig:lbsandoffset} shows example results of the pose-based groom transformation.

\subsection{Groom Deformation Decoder}\label{ssec:gdd}

While the pose-based transformation moves the groom rigidly with the scalp, accounting for global rotation and translation effects, it does not include any deformation as a result of varying external forces resulting from changing body pose and shape. 
The pose-dependent draping effects are accounted for by the groom deformation decoder that produces deformations that, when added to the transformation module's output, produce a physical drape that reacts naturally to gravity and collisions with the underlying body. This module is implemented using a 2D convolutional neural network that takes a vector in $\mathbb{R}^{107}$, which encodes the groom latent code, the body shape parameters and the body pose parameters. 
This input vector is transformed by the network into a texture map of dimension $\mathbb{R}^{T \times T \times N \times 3}$, which encodes the displacements for the vertices along the hair strands. This decoder is trained using a self-supervised loss. Visual results of adding posed based correctives to the rigid groom transformation are shown in Figure~\ref{fig:lbsandoffset}.

\subsection{Physics-Based Self-Supervision}\label{ssec:ep}
\newcommand{\lrest}[0]{l_{\text{rest}}}
We formulate a physics-based loss in order to guide the network to produce physically plausible quasi-static drapes for different grooms under different pose and body variations without the need to pre-compute any simulation data. The proposed loss function is designed to capture the different components that enable the hair to settle in a physically plausible way. This includes the elastic potential of the strands, which limits stretching, bending and twisting of the individual strands, the gravitational pull, the collisions with the underlying body, and the collisions between the hair strands themselves. Additionally, we add a pose regularizer to produce smoothly varying results for smoothly varying skeleton motion sequences. Thus the total consist of the following terms:
\setlength{\abovedisplayskip}{3.5pt}
\setlength{\belowdisplayskip}{3.5pt}
\begin{equation}
\begin{aligned}
 \mathcal{L} &=  \mathcal{L}_\text{elastic\_potential} + \mathcal{L}_\text{gravity} + \mathcal{L}_\text{body\_collision} \\
             &+ \mathcal{L}_\text{self\_collision}  + \mathcal{L}_\text{pose\_reg} 
\end{aligned}
\end{equation}
\noindent\textbf{Elastic Potentials}: 
The elastic potentials allow hair deformation while preserving the length and curvature of each strand. To better preserve the natural curliness of a hairstyle, We first incorporated the Cosserat rod model~\cite{sca2016cosserat} into our system by introducing additional orientation information (stored as unit quaternions $q = (q_0, \textbf{q}^T)^T \in \mathbb{H}$ with the scalar part $q_0$ and the vector part $\textbf{q}$) for each hair segment and optimizing for the additional orientation offsets. Our elastic potential is in the form:
\begin{equation}
\begin{aligned}
\mathcal{L}_\text{elastic\_potential} &= \mathcal{L}_\text{stretch\_shear} + \mathcal{L}_\text{bend\_twist} + \mathcal{L}_\text{unit\_quaternion},
\end{aligned}
\end{equation}
where the Cosserat stretch-shear energy:
\begin{equation}
\mathcal{L}_\text{stretch\_shear} = \frac{1}{2} \sum_\text{edges} \texttt{k}_\text{stretch\_shear} \mathbf{\Gamma}^{T}\mathbf{\Gamma}, 
\end{equation}
is computed from the discrete stretch-shear strain measure $\mathbf{\Gamma} = \frac{\textbf{x}_{i+1} - \textbf{x}_{i}}{\lrest} - \Im(q_{i}\textbf{e}_3\bar{q}_{i})$ for a hair segment $i$, with rest length $\lrest$, end points $\textbf{x}_{i+1}$ and $\textbf{x}_i$, orientation $q_i$ and fixed coordinate basis $\{\textbf{e}_1, \textbf{e}_2, \textbf{e}_3\}$, where $\Im()$ is the imaginary part of the quaternion product. Similarly, the Cosserat bend-twist energy:
\begin{equation}
\mathcal{L}_\text{bend\_twist} = \frac{1}{2} \sum_\text{edge pairs} \texttt{k}_\text{bend\_twist} \mathbf{\Omega}^{T}\mathbf{\Omega},
\end{equation}
is computed from the discrete bend-twist strain measure $\mathbf{\Omega} = \frac{2}{\lrest}(\Im(\bar{q}_iq_{i+1}) - \Im(\bar{q}_i^0q_{i+1}^0))$ for each consecutive hair orientation pair $q_i$ and $q_{i+1}$, with rest orientations $q_{i}^0$ and $q_{i+1}^0$.
An additional unit quaternion constraint is needed for correct bend-twist strain measure, this is enforced for each orientation $q$:
\begin{equation}
\mathcal{L}_\text{unit\_quaternion} = \frac{1}{2} \sum_\text{edges} \texttt{k}_\text{unit\_quaternion} (||q|| - 1)^{2}.
\end{equation}
Although this full Cosserat rod model, which optimizes for both position and orientation offsets, can enable realistic modeling of hair, we observed impractically slow training times and in combination with slow convergence, this approach was deemed infeasible. Instead, we propose an alternative formulation, which optimizes for position offsets only with the following modified Cosserat potential:
\begin{equation}
\mathcal{L}_\text{Cosserat} = \frac{1}{2} \sum_\text{edges} \texttt{k}_\text{Cosserat} \mathbf{\tilde{\Gamma}}^{T}\mathbf{\tilde{\Gamma}},
\end{equation}
where $\mathbf{\tilde{\Gamma}} = \frac{\textbf{x}_{i+1} - \textbf{x}_{i}}{\lrest} - \textbf{d}_3$, with $\textbf{d}_3$ being the unit director along the edge, computed from the rigidly transformed groom.
As the modified Cosserat potential penalizes stretching and shearing simultaneously, we observed that hair would become overly rigid when a large stiffness value $\texttt{k}_\text{Cosserat}$ is used (which is common as hair is almost inextensible). Therefore, to relax $\texttt{k}_\text{Cosserat}$ for more natural hair deformation, we model resistance to stretching explicitly by adding a Hookean potential to penalize the length $l$ of each hair segment to be close to $\lrest$:
\begin{equation}
\mathcal{L}_\text{stretch} = \frac{\texttt{k}_\text{stretch}}{2} \sum_\text{edges}  (l - \lrest)^{2}.
\end{equation}
Alternatively, one can choose to use the potential that penalizes shearing only: $\mathbf{\tilde{\Gamma}} = \frac{\textbf{x}_{i+1} - \textbf{x}_{i}}{l} - \textbf{d}_3$, with $l$ being the segment length. In practice, we observed similar results from both potentials. Our final elastic potential include only two terms and we observe orders of magnitude faster training time when compared with the full Cosserat rod model
\begin{equation}
\mathcal{L}_\text{elastic\_potential} = \mathcal{L}_\text{stretch} + \mathcal{L}_\text{Cosserat}.
\end{equation}

\noindent\textbf{Gravity}:
The gravity potential is modeled using the potential energy of the hair nodes with positions $\mathbf{x}$ as:
\begin{equation}
\mathcal{L}_\text{gravity} = \sum_\text{vertices} -m \mathbf{g}^\top \mathbf{x},
\end{equation}
where $\mathbf{g}$ is the gravitational acceleration and $m$ is the particle point mass.

\noindent\textbf{Body Collision}:
To prevent the hair strands from intersecting with the body geometry, we implement a vertex-triangle based collision model to maintain a minimal distance $D$ between the predicted hair and the outward facing side of the body geometry. 
We exploit the face normal information to ensure the collision response is pointing outwards of the body. The loss is formulated as follows:
\begin{equation}
\mathcal{L}_\text{body\_collision} =  \texttt{k}_\text{bc} \sum_\text{vertices} \text{max} \left( D - d \left( \mathbf{x} \right), 0 \right) ^ 3,
\end{equation}
where $d$ computes the signed distance along the body triangle normal direction and $\texttt{k}_\text{bc}$ is the collision stiffness. Body triangle normals are facing outwards by convention. 

\noindent\textbf{Self-Collision}:
Inspired by work on volumetric methods for hair \cite{bando2003animating, mcadams2009detail, petrovic2005volumetric}, we model hair self-collisions using Smoothed Particle Hydrodynamics (SPH) density estimation. Our energy potential is designed to push vertices apart when they exceed the reference density computed from the rest shape in a smooth and differentiable way: 
\begin{equation}
\mathcal{L}_\text{self\_collision} = \texttt{k}_\text{sc} \sum_\text{vertices} \text{max} \left(\rho(\mathbf{x}) - \rho_\text{rest}, 0 \right)^3,
\end{equation}
% with:
\begin{equation}
\rho(\mathbf{x}) = \sum_j m_j W \left( ||\mathbf{x} - \mathbf{x}_j||, h \right),
\end{equation}
using the smoothing kernel proposed by \citet{bando2003animating}:
\begin{equation}
  W\left(r, h\right) = 
  \begin{cases}
    4 - 6 \left( \frac{r}{h} \right)^2 + 3 \left( \frac{r}{h} \right)^3 & 0 \le r \le h \\
    \left( 2 - \frac{r}{h} \right)^3 & h \le r \le 2h \\
    0 & 2h \le r
  \end{cases}
\end{equation}
where $\rho_\text{rest}$ is the density computed from the rest groom and $h$ is the SPH smoothing length.

\noindent\textbf{Pose Regularization}:
We add a final loss to encourage the groom deformation decoder to predict smoothly varying deformations for smoothly varying input pose signals. During training, for every randomly sampled pose from our motion capture data set, we include $N_\text{pose\_reg}$ continuous frames and penalize the difference with respect to the average deformation in predicted outputs for consecutive frames as: 
\begin{equation}
  \mathcal{L}_\text{pose\_reg} = \texttt{k}_{pr} \sum_{i = 0}^{ N_\text{pose\_reg}}  ||  \bar{\mathbf{x}}_\text{deformation} - \mathbf{x}^i_\text{deformation} ||^2
\end{equation}
where $\bar{\mathbf{x}}_\text{deformation}$ is the average of $\mathbf{x}^i_\text{deformation}$ over all $N_\text{pose\_reg}$ continuous frames per randomly selected pose.

\begin{figure*}[t]
    \centering
    \includegraphics[width=1.0\linewidth]{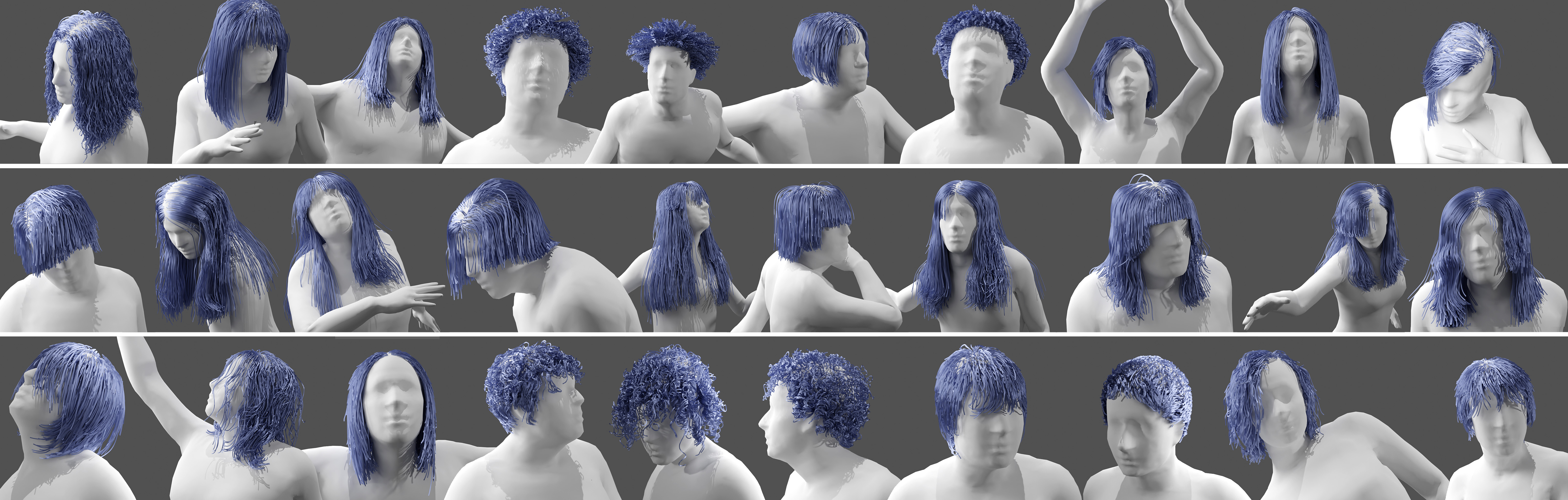}
    \caption{\textbf{Modeling Diverse Hairstyles}: We showcase the versatility of our method in modeling the quasi-static behavior of a plethora of hairstyles and lengths for different body shapes and poses where all results are obtained using the same settings without any manual parameter tuning. Our results cover short/medium/long hair, with various levels of curliness, including straight, wavy, curly, and kinky.}
    \label{fig:groomVariation}
\end{figure*}

\section{Experiments}

\noindent\textbf{Dataset}: We train our networks using the CT-groom data set~\cite{shen2023ct2hair} complemented with additional grooms made by technical artists to showcase a wider variety of hairstyles our method is capable of simulating.

\noindent\textbf{Metrics}: To evaluate our approach we opted for three key metrics: i) the length preservation of the hair strands measured on the segment lengths, ii) the percentage of intersections between the hair and the body, and iii) the preservation of the hair shape measured from the segment orientations.

\noindent\textbf{Baselines}: We compare our method against two optimization-based approaches (Adam, L-BFGS) and one simulation (XPBD~\cite{macklin2016xpbd}) baseline.
We additionally compare to a recent related work GroomGen~\cite{zhou2023groomgen}, which presented a neural quasi-static simulator using supervised training. GroomGen only showed results for varying gravity directions for a single fixed pose and shape, whereas our method generalizes to different shapes and poses. All images show the predicted strands directly without any hair interpolation to provide a fair assessment. When rendering hair for the final application, hair interpolation~\cite{hsu2024real} can be leveraged to obtain the full set of hair strands. 

\noindent\textbf{Implementation Details}:
In all of our results, strands consist of $N = 24$ vertices, which are encoded into $64 \times 64$ texture maps. This choice of texture dimensions provides us with the ability to model several thousands of hair strands, which is sufficient for guide hair simulation. We implement our method with PyTorch~\cite{paszke2017automatic}, with training and inference measurements performed on an AMD Ryzen Threadripper PRO 3975WX CPU and a single NVIDIA RTX A6000 GPU. We refer the reader to the supplementary material for additional details.

\begin{figure}[t]
    \centering
    \includegraphics[width=1.0\linewidth]{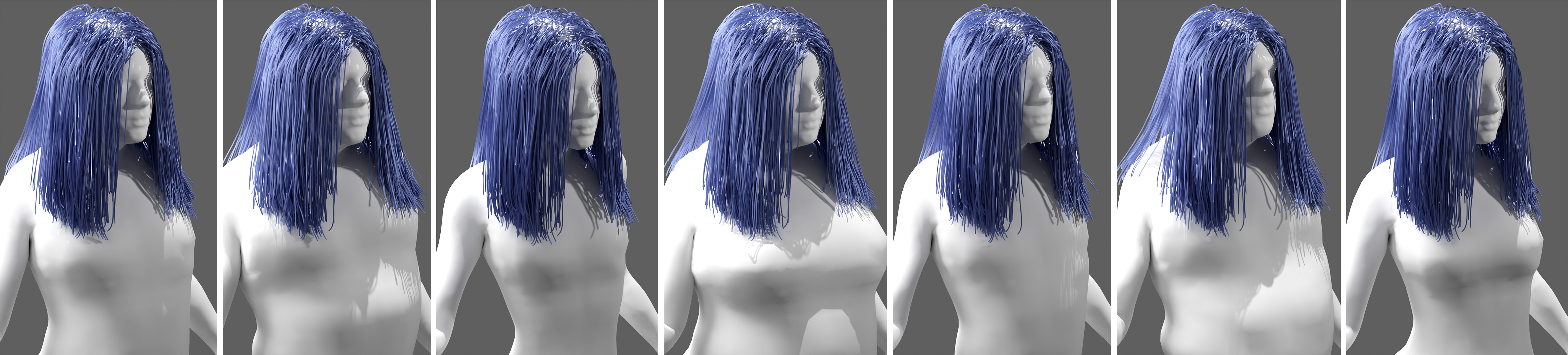}
    \caption{\textbf{Resolving Collisions}: Our method is conditioned on the body shape parameters, enabling it to efficiently resolve collisions with body shape variations. Here we show a static pose and groom under quasi-statically draped under varying body shapes.}
    \label{fig:bodyshapevariation}
    \vspace{-0.15cm}
\end{figure}

\subsection{Qualitative Comparisons}
\noindent\textbf{Body Pose and Shape Generalization}:
Our deformation decoder network is able to ingest body shape parameters directly and effectively learns to adapt the deformations of the hair strands to produce intersection free results for a large variety of body shapes. Figure~\ref{fig:groomVariation} showcases a large sample of pose variations for a diverse set of grooms while in Figure~\ref{fig:bodyshapevariation} we demonstrate body shape generalization.  Implausible deformations are possible when deforming the body in non-physical ways, e.g. when bending the neck backwards beyond limits feasible in real-life and thus absent from our training data set.

\begin{figure}[t]
    \centering
    \includegraphics[width=1.0\linewidth]{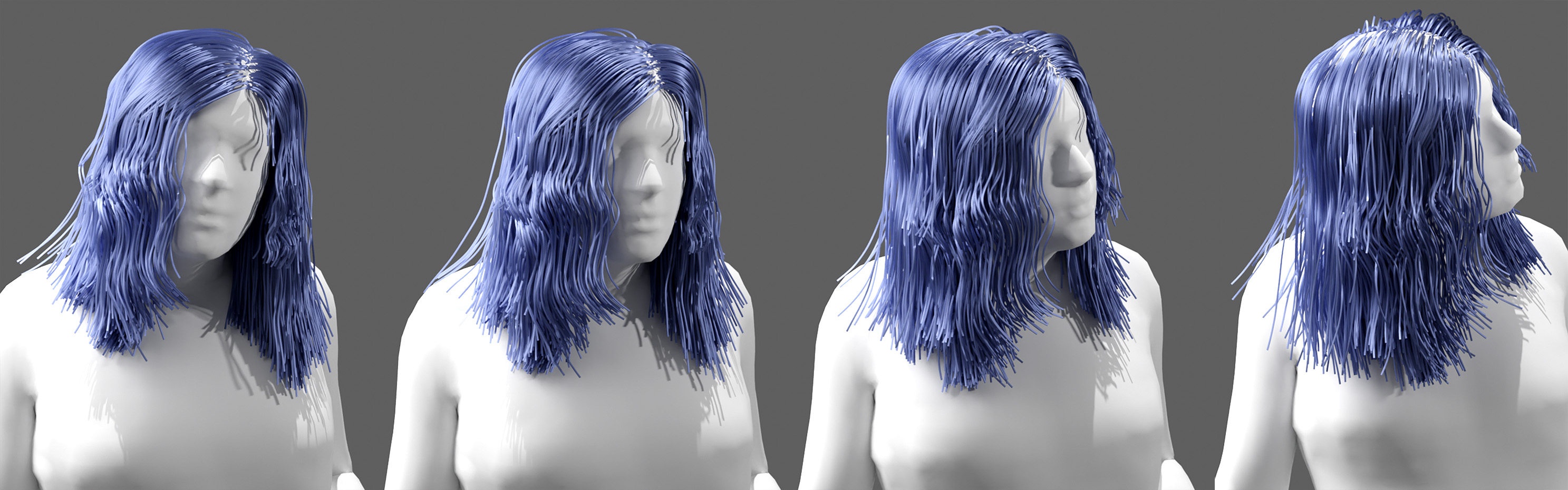}
    \caption{\textbf{Temporal Stability}: To demonstrate how predicted results are smoothly varying with pose changes, we gradually modify the neck rotation. Starting from looking to the right to looking left. Note how the results are smoothly varying where every pose produces a natural drape. Note especially the hair sliding over the shoulders with minimal body intersections.}
    \label{fig:temporalStability}
\end{figure}

\noindent\textbf{Temporal Stability}: Our method demonstrates excellent temporal stability in its predictions when the pose parameters are smoothly varied. This is evident in Figure~\ref{fig:temporalStability}, where the hair smoothly slides over the shoulder as the head turns, avoiding any self-penetration with the body. The supplemental material shows video results.

\noindent\textbf{Groom Generalization}:
We prioritize real-time performance for our approach, therefore opting for a smaller network structure that enables real-time inference results. We show that our model can handle at least 10 distinct grooms. In the supplemental material, we demonstrate that the network already shows some generalization properties for unseen grooms when they are similar to the training set.

\begin{table}[t] 
    \centering
    \caption{\textbf{Quantitative Comparisons}. We compare our method to several optimization and simulation baselines as well as related work. Our method displays comparable metrics to directly optimizing for the positions but does so orders of magnitude faster. Compared to the state-of-the-art method GroomGen, our method produces better quality metrics while being generalizable to body shape and pose, with comparable compute performance, providing the best trade-off between speed and quality. Gravity potentials are provided for reference.}
    \setlength{\tabcolsep}{0.5mm}
    \renewcommand{\arraystretch}{1.2}
    \resizebox{\columnwidth}{!}{
    \begin{tabular}{lrcrrr}
        \toprule
        \multirow{2}{*}{Method}  &$\left(\downarrow \right)$ Time & $\left(\downarrow \right)$ \% body  & $\left(\downarrow \right)$ Length & $\left(\downarrow \right)$ Orientation & Gravity\\
        &   in seconds  & Intersection  &  Preservation  & Preservation & Potential\\  \midrule
        Adam & 179.38 & 0.22 & 103.53 & 76.15 & 3011.33 \\ 
        L-BFGS & 281.18 & 0.22 & 89.53 & 70.22 & 3047.53 \\ 
        XPBD (GPU) \cite{macklin2016xpbd} & 63.26 & 0.01 & 57.96 & 18.10 & 3507.53 \\ \hdashline % 0.009
        GroomGen \cite{zhou2023groomgen} & 0.00249 & 0.39 & 1319.74 & 1281.96 & 3889.78 \\ 
        \textbf{Ours} & 0.00286 & 0.26 & 175.42 & 286.13 & 3399.95 \\
        \bottomrule
    \end{tabular}}
    \label{tab:quantitative}
\end{table}

\subsection{Quantitative Comparisons}

\begin{figure}[t]
    \centering
    \includegraphics[width=0.99\linewidth]{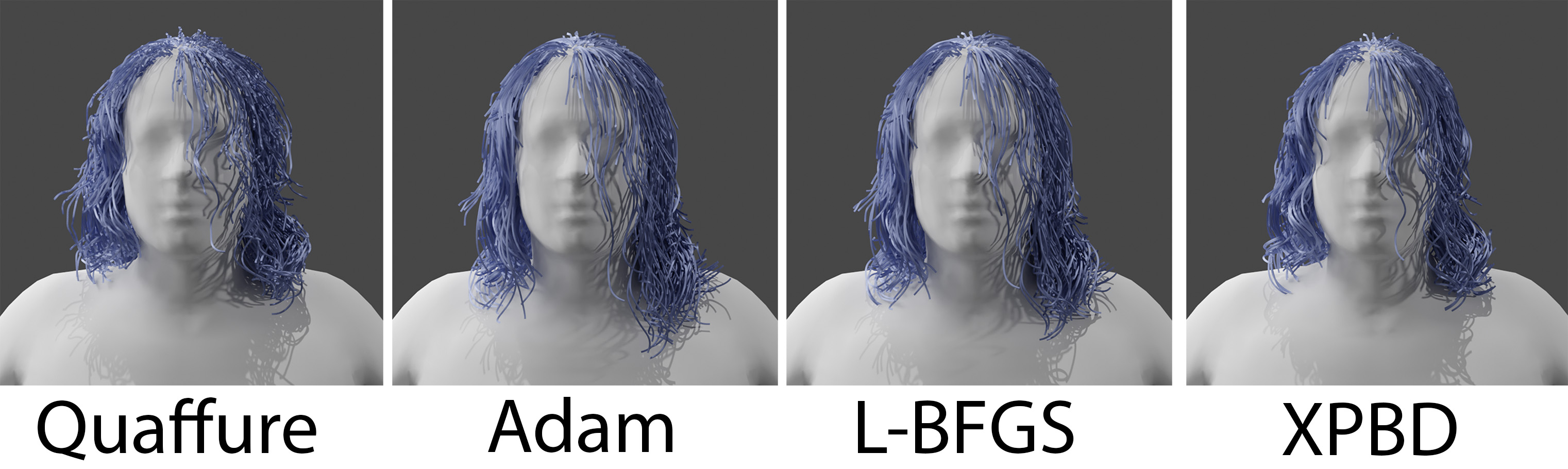}
    \caption{\textbf{Qualitative Comparisons}: Our method produces visually comparable results to directly minimizing the energies or to quasi-static simulation with XPBD while being orders of magnitude faster at inference by memorizing body pose and shape dependent groom deformations.}
    \label{fig:comparison}
\end{figure}

\begin{figure}[t]
    \centering
    \includegraphics[width=1.0\linewidth]{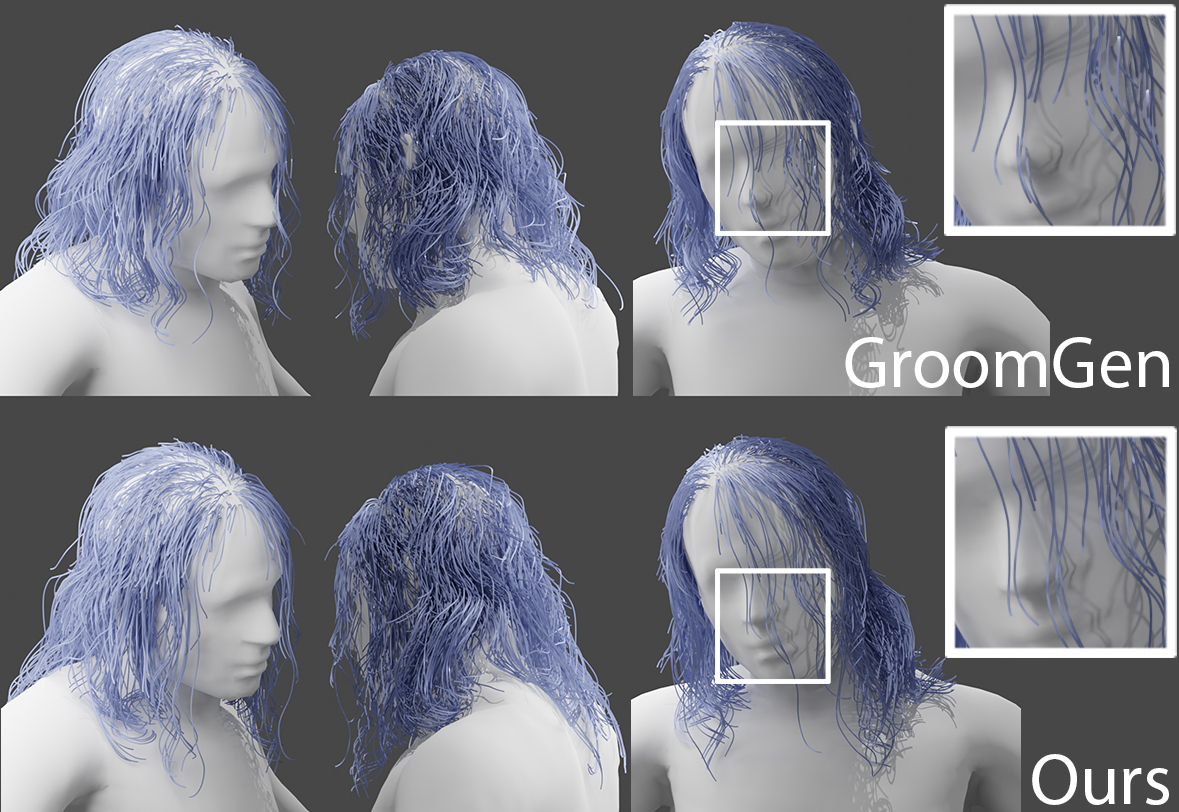}
    \caption{\textbf{Comparisons with GroomGen}: We compare our results (bottom) to those obtained with the neural simulator proposed in GroomGen~\cite{zhou2023groomgen} (top). The inset figure highlights that GroomGen produces results where strands intersect with the head geometry. 
    In contrast, our method produces much fewer body-collisions producing better visual results. See the supplemental video for the full animated comparison.} 
    \label{fig:groomgen}
    \vspace{-0.2cm}
\end{figure}

\noindent\textbf{Comparisons}: The quantitative results are listed in Table~\ref{tab:quantitative}.
As expected, directly modifying the simulated hair positions, either using Adam, L-BFGS or XPBD simulation to minimize the loss produces the highest quality results but at a high computational cost. 
Neural methods like GroomGen and ours are orders of magnitude faster. 
Our method produces significantly better quantitative results when compared to GroomGen at preserving strand lengths as well as the preservation 
of the hair shape measured from the segment orientations. 
Furthermore, it is important to note that, unlike GroomGen, our method handles arbitrary motion and generalizes to body shape. Figure~\ref{fig:groomgen} shows that GroomGen results display frequent intersections of the hair strands with the body. These results demonstrate that our method provides the best trade-offs in terms of performance and quantitative results with the significant added benefit that we do not require any simulated training data. 

\noindent\textbf{Adam \& L-BFGS - Energy Minimization}:
We compare against quasi-static simulation results by minimizing the loss energy with respect to the hair vertex positions directly using Adam and L-BFGS. The results are computed with the highest learning rate that produces stable results. The use of L-BFGS in simulation research has been applied for generating simulations at interactive rates \cite{liu2017quasi}. 
Visual results in Figure~\ref{fig:comparison} show that our method is visually comparable with the quasi-static energy minimization results, but at real-time rates that are independent of groom resolution. This demonstrates that our method is highly effective at learning the mapping that positions deformed hair vertices near the energy minimum.

\noindent\textbf{XPBD - Physics-based Simulation}:
The XPBD method~\cite{macklin2016xpbd} effectively exploits GPU parallelism, which results in fast computation times. Simulation has been the de facto solution to hair animation for years due to the high realism, albeit at a high computational cost. Additionally, time integration of the system can often lead to instabilities, which will result in failures or undesirable results. 
In contrast, our method does not display any time integration instabilities. In our comparison, we model hair using the same energy as described in Subsection~\ref{ssec:ep}. To obtain quasi-static simulation results, we reset the velocities for each time step in the XPBD simulation loop, before applying gravity. We integrate the positions and use $10000$ iterations to minimize the loss. We run 30 time steps to obtain results close to convergence. 

\noindent\textbf{GroomGen}: We additionally compare our method to GroomGen~\cite{zhou2023groomgen}. Code is not publicly available so we compare to our own re-implementation. Instead of training the neural simulator with different gravity directions, we stochastically sample $1000$ body poses with different neck rotations within $30$ degrees along $(x,y,z)$ axes, and run quasi-static hair simulation to obtain the training data. In Figure~\ref{fig:groomgen} provides a qualitative comparison between our approach and GroomGen for a groom that consists of 1919 strands. Both methods provide visually pleasing results but our approach results in significantly fewer intersections of hair with the body. This is highlighted in the inset figure where you can see several hair strands intersect for GroomGen whereas ours remains intersection free.

\subsection{Performance}
\label{sec:performance}

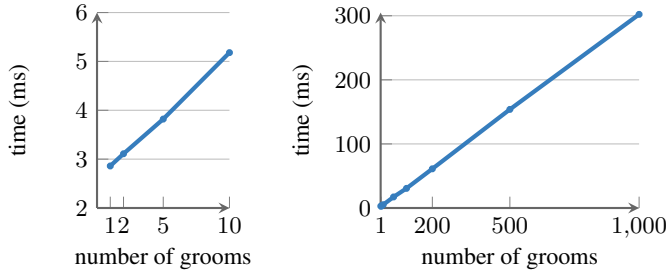
\begin{figure}[t]
    \tikzset{new spy style/.style={spy scope={%
 magnification=5,
 size=1.25cm, 
 connect spies,
 every spy on node/.style={
   rectangle,
   draw,
   },
 every spy in node/.style={
   draw,
   rectangle,
   fill=gray!40,
   }
  }
 }
} 
  
\centering
    \hspace{-0.9cm}
\begin{tikzpicture}[new spy style]
	\begin{axis}[width =0.4\linewidth,
	height = 0.5\linewidth,
    %xmode=log,
    %ymode=log,
    name=ax1,
    legend columns = 3,
    legend cell align=left,
    legend style={font=\footnotesize, at={(0.02, -0.25)}, anchor=north west, line width=0.4pt, draw=.!60!white},
   	xlabel= number of grooms,
        ylabel= time (ms),
        ylabel style={yshift=-0.2cm, font=\small},
        xlabel style={yshift=0.12cm, font=\small},
        axis line style = {thick, color=.!60!white},
		axis y line=left,
	axis x line=bottom,
        ymajorgrids = true,
        tick align=inside,
	tick label style = {font=\small},
        xtick = {1,2, 5, 10},
        xmin = 0.0,
		ymin = 2.0,
		ymax = 6.0],
	\addplot[ultra thick, color=cvprblue, mark=*, mark size=0.5] table [x index=0, y index=1,col sep=space] {Data/performanceUpToTen};
 %\addlegendentry{Ours}
	\end{axis}
 	\begin{axis}[width =0.6\linewidth,
	height = 0.5\linewidth,
        at={($(ax1.south east)+(2cm,0)$)},
    %xmode=log,
    %ymode=log,
    legend columns = 3,
    legend cell align=left,
    legend style={font=\footnotesize, at={(0.02, -0.25)}, anchor=north west, line width=0.4pt, draw=.!60!white},
   	xlabel= number of grooms,
        ylabel= time (ms),
        ylabel style={yshift=-0.2cm, font=\small},
        xlabel style={yshift=0.12cm, font=\small},
        axis line style = {thick, color=.!60!white},
		axis y line=left,
	axis x line=bottom,
        ymajorgrids = true,
        tick align=inside,
	tick label style = {font=\small},
        xtick = {1, 200, 500, 1000},
        xmin = 0.0,
		ymin = 0.0,
		ymax = 305.0],
	\addplot[ultra thick, color=cvprblue, mark=*, mark size=0.5] table [x index=0, y index=1,col sep=space] {Data/performance};
 %\addlegendentry{Ours}
	\end{axis}
%\spy[size=2cm,magnification=3] on (0.0,0.0) in node at (1,1);
\end{tikzpicture}
    \caption{\textbf{Performance \& Scaling}: Our method takes less than 3 ms on average to predict a single groom and the time complexity scales linearly with the number of grooms. The method is able to predict a thousand hairstyles in 0.3 seconds.} 
    \label{fig:performancePlot}
\end{figure}

Predicting a single groom takes $2.86$ ms and predicting a thousand at once requires $0.3$ seconds. The performance is independent of the number of strands due to the fixed size of the representation. Performance for a varying number of groom and pose variations is shown in Figure~\ref{fig:performancePlot}. Strand count for the varying grooms used in training varies between 1500 and 2500. 

\begin{figure}[t]
    \centering
    \includegraphics[width=0.99\linewidth]{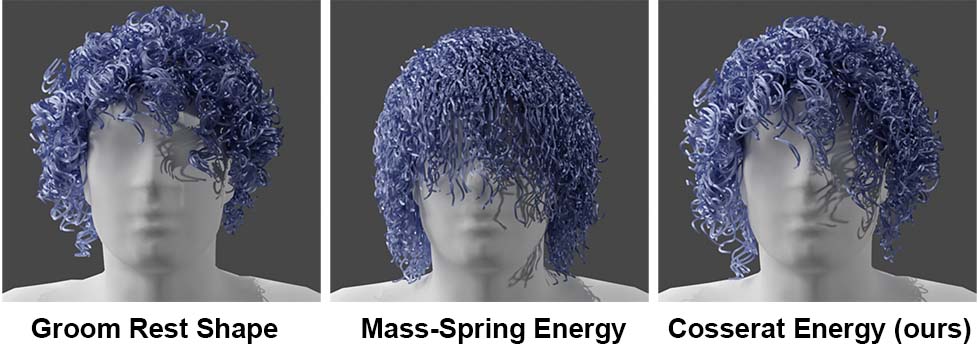}
    \caption{\textbf{Ablation Study}: We perform an ablation of our proposed optimization-friendly Cosserat model against the commonly used mass-spring energy for modeling hair. 
    Our model excels at maintaining the desired hairstyle whereas the mass-spring model struggles with maintaining curliness and volume.} 
    \label{fig:ablation}
\end{figure}

\subsection{Ablations \& Discussion}

\begin{figure}[t]
    \centering
    \includegraphics[width=0.9\linewidth]{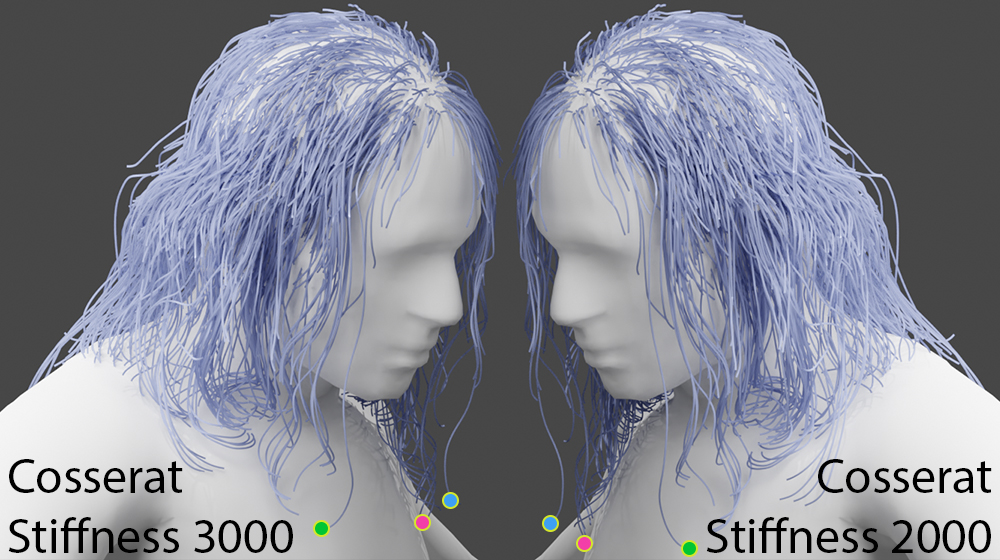}
    \caption{\textbf{Ablation Study on  Cosserat stiffness}: Results obtained by varying the Cosserat stiffness in our proposed physics-based loss that enables artistic control. A higher stiffness results in strands that maintain their rest shape better (left) compared to a lower value (right). The images are mirrored and a few strand hair tip locations are highlighted for an easier comparison.}
    \label{fig:differentCosseratStiffness}
\end{figure}

We perform an ablation experiment on the importance of our proposed Cosserat energy. Figure~\ref{fig:ablation} shows that our model is highly effective at maintaining hairstyle and curliness. We compare our results to those using a mass-spring energy \cite{selle2008mass}. Our model maintains the intended hairstyle well without manual intervention whereas the mass-spring model requires extensive parameter tuning to obtain satisfactory results and the hair still falls mostly flat under gravity. Our model provides artistic control by manipulating the stiffness parameters. Figure~\ref{fig:differentCosseratStiffness} demonstrates the effect of varying material stiffness.

\noindent\textbf{Limitations and Future Work}:
Our pose-based groom transformation module can produce hair where strands are partially initialized inside the body. A better model which results in fewer intersections would make it easier for the model to learn the required deformations. Nonetheless, we observe that our network is able to robustly handle this. Current results use a single set of physical parameters for the hair material properties. This shows that parameter tuning is not required to produce visually pleasing results. However, we plan to condition the network on material parameters to enable artistic control at inference.

\section{Conclusion} 
We present the first self-supervised neural approach to real-time quasi-static hair simulation. Our method enables efficient drape generation for various hairstyles on different body poses and shapes, which varies smoothly with pose variations. Our model leverages a hair deformation decomposition based on a transformation module and a learned deformation decoder, which is trained using self-supervision. We demonstrate complex behavior such as hair sliding over shoulders, maintaining complex hairstyles such as curly grooms and natural draping results due to the effects of gravity. To validate the method, we demonstrate that our decoder is effective and efficient at posing hair strands near the quasi-static energy minimum with better qualitative and quantitative results compared to related work. 

\bibliographystyle{ieeenat_fullname}
\bibliography{References}
\end{document}